\documentclass[lettersize,journal]{IEEEtran}
\usepackage{amsmath,amsfonts}
\usepackage{algorithm}
\usepackage{algpseudocode}
\usepackage{array}
\usepackage[caption=false,font=normalsize,labelfont=sf,textfont=sf]{subfig}
\usepackage{textcomp}
\usepackage{stfloats}
\usepackage{url}
\usepackage{verbatim}
\usepackage{graphicx}
\usepackage{booktabs}
\usepackage{cite}
\usepackage{xcolor}
\usepackage{soul}
\usepackage{multirow} 
\usepackage{fancyhdr}
\usepackage{tabularx}  
\usepackage{makecell} 
\begin{document}

\title{Travel Time and Weather-Aware Traffic Forecasting in a Conformal Graph Neural Network Framework}

\author{Mayur Patil, Qadeer Ahmed, Shawn Midlam-Mohler
\thanks{The authors are with the Department of Mechanical and Aerospace Engineering and the Center for Automotive Research, The Ohio State University, Columbus, OH 43212 USA (e-mail: patil.151@osu.edu; ahmed.358@osu.edu; midlam-mohler.1@osu.edu)}}

\pagestyle{fancy}
\fancyhead{} 
\fancyfoot{} 
\fancyfoot[C]{\color{red}\footnotesize This manuscript has been accepted as a REGULAR PAPER in the Transactions on Intelligent Transportation Systems 2025}
\renewcommand{\headrulewidth}{0pt} 

\maketitle
\thispagestyle{fancy}


\maketitle

\begin{abstract}
Traffic flow forecasting is essential for managing congestion, improving safety, and optimizing various transportation systems. However, it remains a prevailing challenge due to the stochastic nature of urban traffic and environmental factors. Better predictions require models capable of accommodating the traffic variability influenced by multiple dynamic and complex interdependent factors. In this work, we propose a Graph Neural Network (GNN) framework to address the stochasticity by leveraging adaptive adjacency matrices using log-normal distributions and Coefficient of Variation (CV) values to reflect real-world travel time variability. Additionally, weather factors such as temperature, wind speed, and precipitation adjust edge weights and enable GNN to capture evolving spatio-temporal dependencies across traffic stations. This enhancement over the static adjacency matrix allows the model to adapt effectively to traffic stochasticity and changing environmental conditions.
Furthermore, we utilize the Adaptive Conformal Prediction (ACP) framework to provide reliable uncertainty quantification, achieving target coverage while maintaining acceptable prediction intervals. Experimental results demonstrate that the proposed model, in comparison with baseline methods, showed better prediction accuracy and uncertainty bounds. We, then, validate this method by constructing traffic scenarios in SUMO and applying Monte-Carlo simulation to derive a travel time distribution for a Vehicle Under Test (VUT) to reflect real-world variability. The simulated mean travel time of the VUT falls within the intervals defined by INRIX historical data, verifying the model's robustness. 
\end{abstract}


\section{Introduction}
\IEEEPARstart{T}{raffic} congestion in cities is a major problem that greatly affects economic productivity, environmental sustainability, and public safety. Based on the INRIX Global Traffic Scorecard for 2023 \cite{INRIX}, drivers in the United States wasted an average of 42 hours each year, resulting in a total cost of \$70 billion for the country, or \$733 per driver. In contrast to 2022, this represents a 4-hour uptick in delays and underscores a 15\% surge in related economic expenses. Despite 40\% of U.S. urban areas reaching or surpassing pre-COVID traffic levels, most are still dealing with increased traffic congestion, as 78\% of urban areas are seeing a surge as compared to the previous year. The increase in midday travel after COVID, showing a 23\% rise from 2019, indicates changes in commuting and underscores the importance of accurate traffic prediction models to support smart transportation systems and reduce traffic jams.

Precise traffic state predictions enable transportation managers to make informed decisions such as optimizing traffic lights, efficient routing, and issuing real-time traffic alerts. However, forecasting traffic states in an urban setting is a complex task due to the highly stochastic nature of traffic dynamics. The complexity is caused by factors like uncertain travel times, varying weather conditions, availability of data, fatal crashes along the road, and various other issues. This added layer of complexity poses an onerous task that needs to be solved by any prediction model. 

In recent years, Graph Neural Networks (GNNs) have gained a lot of traction for modeling spatial and temporal data for traffic count stations in a road network \cite{STGCN, DCRNN, TGCN, ASTGCN, Wavenet, GMAN}. It represents the information in terms of nodes for stations and edges for connected roads which makes them a deserving candidate for traffic state prediction tasks. These models are capable of integrating spatial dependencies between traffic nodes and their temporal patterns as well. Again, despite their advantages, these GNN models utilize static adjacency matrices to consume the connections coming from the neighboring nodes in a traffic network. This static representation of information neglects the dynamic behavior of the vehicles which can arise due to fluctuating travel times and other environmental factors. 
\begin{figure}[!b]
    \centering
    \includegraphics[width=\linewidth]{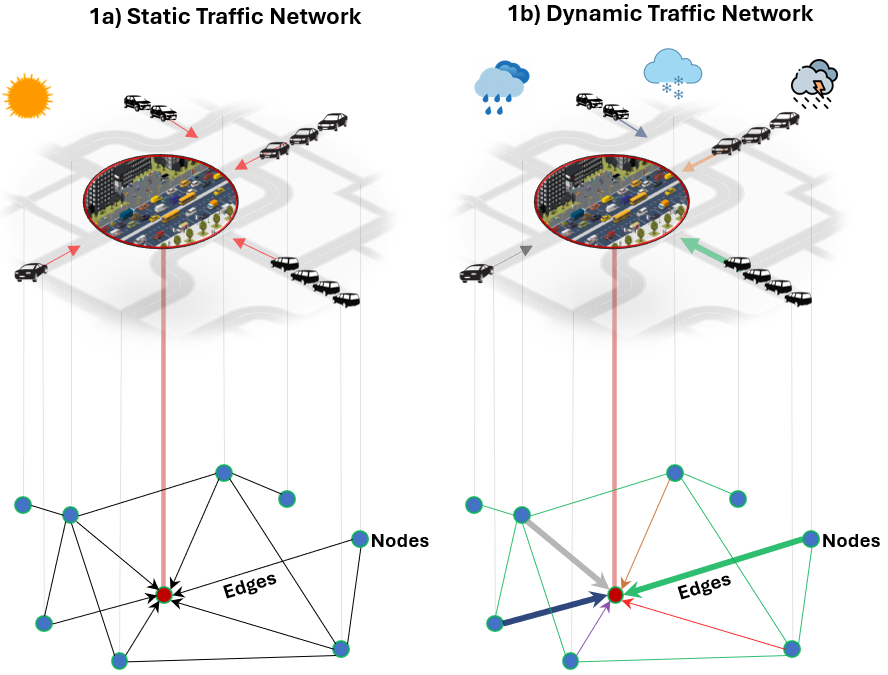} 
    \caption{Comparison of static and dynamic traffic network models}
    \label{fig:intro}
\end{figure}
Figure \ref{fig:intro} illustrates the difference between traditional static traffic network models and the proposed dynamic traffic network. The static network (Figure 1a) relies on fixed edge weights, which fail to account for the dynamic nature of real-world traffic conditions. In contrast, the dynamic network (Figure 1b) adapts its edge weights in real-time based on varying travel times between nearby nodes and a central focused node considering external factors like weather conditions (as represented by varying weights and colors of connected edges and traffic flow). Additionally, it incorporates data availability from the stations with high data richness and others with sparse data coverage to ensure the network leverages a comprehensive view of the traffic network.

In our previous work \cite{Patil}, we focused on some of these issues by leveraging a GNN-based model integrated with Adaptive Conformal Prediction (ACP) for uncertainty quantification. We also combined the data from Continuous Count Stations (CCS) and Non-Continuous Count Stations (N-CCS) to exploit information/insights from rich as well as sparse traffic flow data respectively. Although our approach showed acceptable results, we as well, did not consider the dynamic interactions between the traffic count stations nor the impact of environmental conditions on traffic flow. Therefore, based on our initial work, this paper introduces the extension to our original framework enhancing the urban traffic flow prediction by incorporating adaptive adjacency matrices into a GNN model. These matrices are constructed using real-world travel time distributions between traffic stations modeled as log-normal distributions to reflect the inherent variability in urban traffic. By considering different Coefficient of Variation (CV) values in travel time distribution, the model adapts to different traffic scenarios ranging from low to heavy traffic conditions.

Environmental factors especially weather play a critical role in influencing traffic flow. Inclement weather conditions can lower road capacity, slow traffic, and increase the chances of vehicle crashes. As per the Federal Highway Administration report \cite{NHTSA2016}, there are more than 5,891,000 vehicle crashes each year, with approximately 21\% of these crashes - about 1,235,000 - being weather-related. On average, almost 5,000 individuals die and over 418,000 people are injured in weather-related crashes yearly. The record shows that 70\% of weather-related crashes happen on wet pavement, and  46\% during rainfall. While 18\%  crashes occur during snow or sleet, 13\% occur on icy roads, 16\% due to snowy or slushy roads, and only 3\% occur in foggy conditions. To tackle this issue, we integrate weather data from Road Weather Information System (RWIS) sensors within Ohio \cite{RWIS} that have records for temperature, wind speed, and precipitation directly into the dynamic adjacency matrices. This development allows the model to adjust edge weights dynamically according to environmental conditions and improves its ability to predict traffic flow under varying weather conditions.  

We kept the aspect of augmenting data availability scores from CCS and N-CCS stations in the adjacency matrices unchanged as our prior study \cite{Patil} proposed. By including the data ``type", we just make sure of comprehensive network coverage. Basically, we did not overlook the sparse N-CCS data, which, despite its limitations, provides valuable insights into traffic flow.

In order to perform validation, we conducted experiments using actual traffic data from Columbus, Ohio. We replicate/simulate traffic scenarios in the Simulation of Urban Mobility (SUMO) environment and run Monte Carlo simulations to analyze travel time distributions for a Vehicle Under Test (VUT), which was then compared with historical data from INRIX. This step demonstrated the model's exhaustive prediction robustness and applicability. To revisit the main contributions of the paper:
\begin{enumerate}
    \item We develop a stochastic method for constructing adaptive adjacency matrices that reflect real-time travel time variability between traffic stations using log-normal distributions and incorporating CV.

    \item We improve the prediction capability of the model by integrating weather factors into the adaptive adjacency matrices. 

    \item We incorporate data availability scores from CCS and N-CCS stations into the adjacency matrices to prioritize reliable data sources.

    \item We validate our proposed framework by applying Monte-Carlo simulations in SUMO to simulate different traffic scenarios to compare travel time distributions for a VUT with INRIX travel time data.
\end{enumerate}

The remainder of this paper is organized as follows. Section II reviews related work in traffic flow prediction. Section III details the methodology of our proposed framework. Section IV compares the experiments and results to other baseline models and finally, Section V concludes the paper and suggests directions for future research.

\section{Related Work}
\subsection{Traditional Physics-Based Models for Traffic Flow}
Physics-based (process-dynamics) models came before today’s data-driven methods and are still widely used for short-term traffic prediction and control. They update traffic states using basic rules from traffic flow theory - flow conservation, considering road capacity, and relating flow, speed, and density. Some representative models include the Cell Transmission Model (CTM) \cite{Daganzo94}, Link Transmission Model (LTM) \cite{Yperman07}, and store-and-forward (SAF, link-queue) formulations for urban networks \cite{Aboudolas2009}. At the PDE (partial differential equation) level, the Lighthill-Whitham-Richards (LWR) theory \cite{LWR55} and higher-order variants such as Payne–Whitham (PW) \cite{Payne1971} and Aw-Rascle-Zhang (ARZ) \cite{AwRascle2000} encode similar ideas in continuous form. In practice, these models are often paired with Kalman-filter families for state estimation and parameter calibration. However, they require careful tuning and may struggle under dynamic changes (e.g., incidents, work zones, weather) \cite{OJITS2022}. These limits motivate the data-driven spatio-temporal methods reviewed next.

\subsection{Spatio-Temporal Models for Traffic Prediction}
Accurate traffic state prediction is decisive for these days Intelligent Transportation Systems (ITS) that bolsters traffic management, reduce congestion, and enhance road safety. Over the years, various data-driven models have been proposed to predict traffic states addressing the limitations of physics-based models. These range from traditional statistical methods to advanced spatio-temporal deep learning techniques. Traditional time-series models include ``Autoregressive" (AR) models like Autoregressive Integrated Moving Average (ARIMA), and ``smoothening" models like Simple Exponential Smoothing (SES), which have been widely used for their simplicity and effectiveness in analyzing temporal/sequential patterns \cite{BoxJenkins, SES, Shuvo2021, Lippi2013}. However, these models fail to capture the spatial dependencies in transportation networks as they treat the traffic sensors independently. Therefore, the limitation reduces their effectiveness in traffic prediction tasks. To address these shortcomings, researchers have turned to the Machine Learning (ML) field and deep learning approaches that can model both spatial and temporal dependencies. To begin with, Recurrent Neural Networks (RNNs) and Long Short-Term Memory (LSTM) models have been applied to traffic prediction tasks that showed improved performance by capturing temporal dependencies in sequential data \cite{Fu2016, Tian2015, Ma2015}. Anyhow, these models still lack the capability to model spatial relationships among different traffic station sensors explicitly. 

The pioneering works in this area started with the development of the Spatio-Temporal Graph Convolutional Network (STGCN) by Yu \textit{et al.} \cite{STGCN} that used graph convolutions for handling spatial dependencies with temporal convolutions for handling temporal dynamics. Later, Li \textit{et al.} introduced the Diffusion Convolutional Recurrent Neural Network (DCRNN) that models traffic flow as a diffusion process on a directed graph and used RNNs to capture temporal dependencies \cite{DCRNN}. On the same lines, Zhao \textit{et al.} proposed the Temporal Graph Convolutional Network (T-GCN) using GCNs with gated recurrent units for spatio-temporal modeling \cite{TGCN}. So, other models based upon these foundational works were researched to include attention mechanisms, intricate dynamic graph structures, and comprehensive temporal processing to improve prediction accuracy. Guo \textit{et al.} developed the Attention-Based Spatial-Temporal Graph Convolutional Network (ASTGCN) that used attention mechanisms to capture dynamic dependencies \cite{ASTGCN}. Wu \textit{et al.} proposed Graph WaveNet that integrates adaptive adjacency matrices and causal convolutions to capture some long-term temporal dependencies \cite{Wavenet}. Zheng \textit{et al.} introduced the GMAN model by utilizing an encoder-decoder architecture with spatial and temporal attention to predict traffic flow \cite{GMAN}. However, these models utilized static connections to represent spatial information in a traffic network that overlooks the dynamics of the ever-changing traffic.

Intelligent transportation research is constantly progressing toward vehicle-centric solutions, such as cooperative unmanned-vehicle fleets for real-time sensing and control, in addition to network-level forecasts \cite{UAV}. Recent research demonstrates that distributed machine-learning agents in conjunction with LoRa gateways can provide low-power connectivity for green ITS services at the communication layer \cite{LoRa}. Our study is correlational because accurate macro-scale forecasts are still necessary for both components.

\subsection{Adaptive Adjacency Matrices}
Despite the advancements, many GNN-based models still rely on static adjacency matrices constructed based on predefined criteria such as physical distances (Euclidean or non-Euclidean) or connectivity (traffic flow or density). These static matrices do not adapt to the dynamic nature of traffic networks, where node relationships can change due to varying traffic conditions, incidents, or roadworks. So, to address the limitations of static adjacency matrices, studies have explored the use of adaptive adjacency matrices that can capture the evolving changes in a traffic network. Wu \textit{et al.} propose Graph WaveNet, integrating adaptive adjacency matrices that learn from the data, allowing the model to discover hidden spatial dependencies \cite{Wavenet}. Another study by Bai \textit{et al.} proposed the Adaptive Graph Convolutional Recurrent Network (AGCRN) which learns node-specific pattern relationships and provides a flexible representation of spatial factors \cite{AGCRN}. Ta \textit{et al.} work developed a model that combined adaptive graph structure with an attention mechanism to capture spatio-temporal dependencies \cite{Ta2022}. Zhang \textit{et al.} introduced AdapGL, an adaptive graph learning algorithm that incorporates a novel graph learning module within a spatio-temporal neural network \cite{AdapGL}, which adaptively captures complex node dependencies during training by optimizing the adjacency matrix using an expectation-maximization approach. Cui \textit{et al.} proposed ADSTGCN, a Dynamic Adaptive Deeper Spatio-Temporal Graph Convolutional Network for multi-step traffic forecasting \cite{ADSTGCN} and addressed the over-smoothing issue in deep graph convolutional networks (GCNs) by employing dynamic hidden layer connections and adaptively adjusting hidden layer weights, which allows the model to capture unknown dynamic changes in the traffic network. Xu \textit{et al.} applied an adaptive adjacency matrix-based GCN to taxi demand forecasting \cite{Xu2022} by utilizing a two-stage clustering approach to discover virtual stations based on geographical and demand characteristics and combined non-negative matrix factorization with GCNs to extract node features and compute adaptive adjacency matrices. Together, these studies demonstrate why adaptive graphs matter, and we build on that idea by refreshing the adjacency matrix with travel-time samples and weather information, so the graph keeps pace with real-time road conditions. 

\subsection{Weather-Integrated Adaptive Adjacency Matrices}
In addition to adaptive adjacency matrices, stochastic factors such as weather also play a major role in connectivity, because traffic depends heavily on road conditions and visibility. Studies have shown that adverse weather conditions can lead to increased congestion, travel times, and crash rates \cite{Cools2010}, \cite{Agarwal2005}. Some researchers have incorporated the weather data as additional inputs in traffic prediction models. Qi \textit{et al.} proposed a Spatial-Temporal Fusion Graph Convolutional Network (STFGCN) that combined weather attributes with traffic data to predict traffic flow \cite{Qi2024}. Their hybrid deep learning model also integrates multiple other features, demonstrating improved performance over models that considered only normal conditions. The study conducted by Chen \textit{et al.} introduced Attentive Attributed Recurrent Graph Neural Network (AARGNN) to consider multiple dynamic factors in traffic flow prediction \cite{AARGNN}. It constructed an attributed graph where various static and dynamic factors such as spatial/semantic distances, road characteristics/events, and other global contexts into node attributes. A study by Alourni \textit{et al.} proposed a hybrid deep learning model, Double Attention Graph Neural Network BiLSTM (DAGNBL) that used correlated weather data in the forecasting model along with spatial dependencies \cite{BiLSTM}. Their approach improved prediction accuracy and achieved a mean absolute percentage error of 5.21\% showing a better capability while integrating weather data into the model. Following, Al-Selwi \textit{et al.} presented an investigation of the impact of weather data on various traffic prediction models \cite{AlSelwi2022}. In their findings, they also highlighted how most existing models overlook the weather factors and their impact on prediction accuracy. So, we can clearly see how the evolving recognition of weather factors is a critical component in traffic prediction models and their integration through feature aggregation, attributed graphs, or via attention mechanisms. However, integrating weather information directly into the graph structure of a GNN model explicitly in the dynamics of an adjacency matrix remains an unexplored avenue. 

\subsection{Travel Time-Based Adaptive Adjacency Matrices}
\begin{figure}[h]
    \centering
    \includegraphics[width=\linewidth]{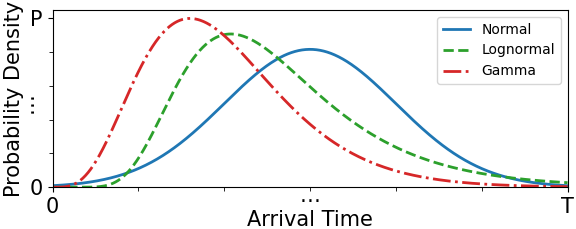} 
    \caption{Comparison of Arrival Time Distributions}
    \label{fig:distributions}
\end{figure}

Building upon the concept of adaptive adjacency matrices and modeling travel time variability, we connect this back to our previous work \cite{Patil}, where we initially proposed using travel times between traffic stations to construct the adjacency matrix. However, in that model, the adjacency matrix remained static and did not account for the stochastic nature of travel times under different traffic conditions. In this paper, we build on the same method by introducing adaptive adjacency matrices using a log-normal distribution to represent travel time. By adjusting the CV values, we are able to present various traffic situations, from low to heavy traffic scenarios.
Numerous researches have shown that travel times in transportation networks commonly exhibit log-normal distributions because of their right-skewed and non-negative features. Shen \textit{et al.} \cite{Shen2019} and colleagues introduced a composite travel time model utilizing distribution fitting techniques with Automatic Vehicle Location (AVL) data, determining that log-normal distributions were a suitable fit across different traffic scenarios. Mazloumi \textit{et al.} \cite{Mazloumi2010} in his study analyzed the travel times in public transportation on a daily basis by utilizing vehicle GPS data and concluded that peak hour travel times usually conform to a normal distribution fit, whereas the non-peak travel times showed a log-normal distribution. Yan \textit{et al.} \cite{Yan2012} created a robust optimization model for designing bus schedules that considers travel time uncertainty by representing them using a log-normal distribution. Moreover, many studies mentioned in Dessouky's review \textit{et al.} \cite{Dessouky1999} have used skewed distributions like log-normal or gamma for modeling bus arrival times showcasing their effectiveness in capturing travel time variability. For example, Turnquist used log-normal distributions for arrival times \cite{Turnquist1978}, and Strathman and Hopper did the same in their study \cite{Strathman1993}, while Guenthner and Hamat used gamma distributions \cite{Guenthner1988}. Seneviratne \cite{Senevirante1990} studied travel time variation using a normal and gamma distribution, concluding the importance of selecting the correct distribution based on real-world data. Figure \ref{fig:distributions} shows a sample Normal, Lognormal, and Gamma distribution. The Normal distribution models have a stable variability, whereas the Lognormal and Gamma distributions are right-skewed and are appropriate for non-negative, time-to-event data.
\vspace{-5pt}
\vspace{-15pt}

\subsection{Uncertainty Quantification}

Incorporating uncertainty quantification into traffic flow prediction models is crucial due to the inherent unpredictability of traffic conditions caused by uncertain accidents, weather, and traffic volumes. Traditional methods for uncertainty quantification include Bayesian inference, quantile regression, Monte Carlo dropout, deep ensembles, and bootstrap aggregation \cite{Bayesian, QR, DE, BA, Dropout}. These techniques provide a measure of confidence in predictions but often require strong assumptions about data distributions or involve substantial computational complexity. More recently, Conformal Prediction (CP) and its extension Conformalized Quantile Regression (CQR) have provided distribution-free calibration rules \cite{CP, CQR}. CP offers valid prediction intervals without any distributional assumption \cite{AdapCP}, updating them with residuals to maintain coverage under non-stationarity, yet its bands can widen if the forecaster underfits. Whereas, CQR narrows these bands by learning two extra quantile heads and then calibrating them, which increases training cost. Its coverage can still degrade when traffic patterns shift drastically. In this work, we continue using ACP \cite{Patil} integration method for uncertainty quantification. ACP is a variation coming off of CP capable of providing an efficient and robust way to quantify uncertainty bounds by directly utilizing the residuals from past predicted values.
In comparison with traditional methods, ACP provides benefits in traceability and adaptability, making it suitable for real-time traffic prediction applications.

\section{Proposed Methodology}
\subsection{Graph Formulation and Data Availability Matrix}
Following our prior methodology \cite{Patil}, we model the urban traffic network as a directed graph $\mathcal{G} = (\mathcal{V}, \mathcal{E})$, where $\mathcal{V}$ represents the set of nodes corresponding to traffic count stations, and $\mathcal{E}$ represents the set of directed edges corresponding to road segments connecting these stations. Each node $i \in \mathcal{V}$ can be either a CCS or N-CCS, and these nodes are assigned a data availability score to reflect the varying reliability of traffic count data. For CCS stations, which have full data availability, the score is set to 1. For N-CCS stations, the score is calculated based on the available count data $C_i$:

\begin{equation}
A_i =
\begin{cases}
1, & \text{if node } i \text{ is CCS}\\[4pt]
\displaystyle\frac{C_i}{\max_j C_j}, & \text{if node } i \text{ is N‑CCS}
\end{cases}
\end{equation}%
where $C_i$ is the count data from N-CCS node $i$, and $\max_{j} (C_j)$ is the maximum count observed among all N-CCS stations. This normalization ensures that stations with higher data availability have higher scores. Next, we combine these individual scores into a data availability matrix $\mathbf{A}_{\text{avail}}$, which captures the joint reliability of data between pairs of stations:

\begin{equation} 
\mathbf{A}_{\text{avail}}[i,j] = A_i \cdot A_j
\end{equation}
where $i$, $j$ $\in \mathcal{V}$. $\mathbf{A}_{\text{avail}}[i,j]$ thus encodes the joint reliability of the edge $(i,j)$.

\subsection{Adaptive Adjacency Matrix with Travel Time}
We introduce an adaptive adjacency matrix $\mathbf{A}_{\text{adaptive}}$ to capture traffic flow dynamics. This matrix adjusts based on travel time variability and weather conditions, thus providing a realistic representation of traffic conditions.
Travel times between traffic stations are inherently variable due to factors such as traffic congestion, accidents, and roadworks. We model travel time variability between stations $i$ and $j$ using a log-normal distribution. This distribution is appropriate because travel time data typically exhibits positive-valued and right-skewed characteristics. The mean travel times between stations are represented by the static matrix $\mathbf{T}_{\text{mean}}$, which can be obtained from historical average travel times or estimated based on the actual distances and average speed limits between stations. The smallest observed link travel time was 1.65 min for our sensor locations, and if any rare reading near zero is encountered, it will be treated as missing and replaced by the link’s historical median for that time of day.
To quantify travel time variability explicitly, we introduce the CV variation, defined as the ratio of the standard deviation to the mean travel time and to categorize traffic scenarios according to their variability:

\begin{equation} \text{CV} = {\sigma}/{\mu} \end{equation}
where $\sigma$ is the standard deviation of the travel time, and $\mu$ is the mean travel time from $\mathbf{T}_{\text{mean}}$. We evaluate model performance under five traffic scenarios, each characterized by different CV values \((\text{CV} \in \{0.1, 0.3, 0.5, 0.7, 1.0\})\), where higher CV implies greater variability in travel time. These values span the range observed on our corridor-smaller CVs (e.g., 0.1-0.3) reflect more stable periods such as late night with tighter spreads, whereas larger CVs (e.g., 0.7-1.0) align with peak periods where travel times fluctuate more widely. For each scenario, we compute the parameters of the log-normal distribution as:

\begin{equation}
\sigma_{\text{ln}} = \sqrt{\ln(\text{CV}^2 + 1)}
\end{equation}
\vspace{-2mm}
\begin{equation}
\mu_{\text{ln}} = \ln(\mu) - \frac{\sigma_{\text{ln}}^2}{2}
\end{equation}
\noindent
where $\sigma_{\text{ln}}$ and $\mu_{\text{ln}}$ are the parameters of the log-normal distribution used to sample stochastic travel times for each traffic scenario. Using the computed parameters, we then generate travel time matrix adaptive to traffic dynamics, $\mathbf{T}_{\text{dynamic}}$, by sampling 50 instances from the log-normal distribution for each traffic scenario:

\begin{equation} T^{\text{dynamic}}_{ij} \sim \text{LogNormal}(\mu_{\text{ln}}, \sigma_{\text{ln}}) \end{equation}

\subsection{Integrating Weather Influence into the Adjacency Matrix}
We use weather data from 215 Ohio RWIS (O-RWIS) sensors to adjust edge weights in the adjacency matrix. Weather variables are mapped to 273 traffic count stations (225 CCS and 48 N-CCS with $\geq$ 40\% data availability) using Inverse Distance Weighting (IDW). Categorical precipitation types (e.g., heavy snow, moderate rain, none) are numerically encoded for consistency. The IDW enables estimation of weather conditions at stations lacking direct sensor measurements by weighting nearby sensor data inversely to distance.

Let $S = { s_1, s_2, \dots, s_N }$ denote the set of O-RWIS sensors located at coordinates $(x_k, y_k)$ and providing weather observations $w_k$ (temperature, wind speed, and precipitation type in our case). Let $T = { t_1, t_2, \dots, t_M }$ represent the set of traffic count stations at locations $(x_i, y_i)$. For each traffic station $t_i$, the interpolated weather variable $\tilde{w}_i$ is computed using IDW:
\vspace{1.2em}
\begin{equation} \tilde{w}_i = \frac{\sum_{k=1}^{K} w_k \cdot \frac{1}{(d_{ik} + \epsilon)^p}}{\sum_{k=1}^{K} \frac{1}{(d_{ik} + \epsilon)^p}} \end{equation}
where, $d_{ik}$ is the geodesic distance between traffic station $t_i$ and O-RWIS sensor $s_k$, $p$ is the power parameter controlling the influence of distance (commonly set to $p=1$ or $2$), $\epsilon$ is a small constant to avoid division by zero, and $K$ is the number of nearest O-RWIS sensors considered (we use $K=3$). Algorithm 1 shows the pseudo-code for sensor mapping.

\begin{algorithm}[H]
\caption{Weather Data Mapping using IDW}\label{alg:alg1}
\begin{algorithmic}
\State \textbf{Input:} $S = \{s_1, \dots, s_N\}$, $T = \{t_1, \dots, t_M\}$, $K$, $p$, $\epsilon$
\State \textbf{Output:} $\tilde{w}_i$: Interpolated weather variable for each $t_i$
\State
\For{$t_i \in T$}
    \State Compute $d_{ik} = \text{geodesic}((x_i, y_i), (x_k, y_k)) \; \forall \; s_k \in S$
    \State Find $K$ nearest sensors: $\mathcal{N}_i = \text{argsort}(d_{ik})[:K]$
    \For{each weather variable $w_k$}\\
        \State $\tilde{w}_i = \dfrac{\sum_{k \in \mathcal{N}_i} w_k / (d_{ik} + \epsilon)^p}{\sum_{k \in \mathcal{N}_i} 1 / (d_{ik} + \epsilon)^p}$
    \EndFor
\EndFor\\
\Return $\{\tilde{w}_i\}_{i=1}^M$
\end{algorithmic}
\end{algorithm}

\noindent
Next, we integrate the interpolated weather variables into the sampled adjacency matrix by calculating correlations between \(T_{ij}^\text{dynamic}\) and weather features namely temperature (\(\rho_{ij}^\text{temp}\)), wind speed (\(\rho_{ij}^\text{wind}\)), and precipitation (\(\rho_{ij}^\text{precip}\)). Each correlation coefficient is computed as:
\begin{equation}
\rho_{ij}^\text{weather} = \text{corr}\left(T_{ij}^\text{dynamic}, \frac{w_i + w_j}{2}\right)
\end{equation}
where \(w_i\) and \(w_j\) are the time series of weather features at stations \(i\) and \(j\), respectively. The adjusted matrix producing a weather-aware travel-time matrix is then expressed as:
\begin{equation}
T_{ij}^\text{adjusted} = T_{ij}^\text{dynamic} \cdot \left(1 + \sum_{k} \alpha_k \cdot \rho_{ij}^k \right)
\end{equation}
where the weights \(\alpha_k\) (\(k \in \{\text{temp, wind, precip}\}\)) are calculated via Equation (10-11) by fitting a regression model between each weather feature and its travel time value in the matrix:

\begin{equation}
T_{ij}^\text{dynamic} = \sum_{k} \beta_k \cdot \rho_{ij}^k + \epsilon
\end{equation}
where \(\beta_k\) are the regression coefficients representing the influence between each weather feature and travel time value, and \(\epsilon\) is the error term. Also, we normalize the weights \(\alpha_k\) as:

\begin{equation}
\alpha_k = \frac{|\beta_k|}{\sum_{k} |\beta_k|}.
\end{equation}
The dynamic adjacency matrix is computed using a Gaussian kernel transformation \cite{STGCN}:
\begin{equation}
\mathbf{A}_{\text{dynamic}}[i,j] = \exp\left(-\frac{\left(T_{ij}^\text{adjusted} / T_{\max}\right)^2}{2\sigma^2}\right)
\end{equation}
where \(T_{\max}\) is the maximum travel time, and \(\sigma^2\) controls the spread of the Gaussian kernel. Finally, we obtain the adaptive adjacency matrix, which merges the weather‑adjusted dynamic connectivity with node‑level data reliability:
\begin{equation}
\mathbf{A}_{\text{adaptive}}[i,j] = \mathbf{A}_{\text{dynamic}}[i,j] \cdot \mathbf{A}_{\text{avail}}[i,j],
\end{equation}
thereby encoding both spatio‑temporal interactions and data‑quality weighting in a single graph structure for the GNN.

\subsection{Model Architecture and Adaptive Conformal Prediction (ACP)}
We first emply a Graph Attention Networks (GATs) to capture spatial dependencies among nodes in the traffic graph. Each node aggregates information from its neighbors, weighted by an attention mechanism that emphasizes more relevant spatial features. The following equations describe how attention scores and node representations are computed:

\begin{equation}
    \mathbf{h}_i = \sigma\left( \sum_{j \in \mathcal{N}(i)} \alpha_{ij} \mathbf{W} \mathbf{F}_j \right)
\end{equation}

\begin{equation}
    \alpha_{ij} = \text{softmax}_j\left( e_{ij} \cdot \mathbf{A}_{\text{adaptive}}[i,j] \right)
\end{equation}

\begin{equation}
    e_{ij} = \text{LeakyReLU}\left( \mathbf{a}^\top \cdot \left[ \mathbf{W} \mathbf{F}_i \| \mathbf{W} \mathbf{F}_j \right] \right)
\end{equation}
where $F_i, F_j$ are the input feature vector of node $i, j$ respectively, $h_i$ is the hidden representation, $\mathcal{N}(i)$ denotes the neighboring nodes, $\alpha_{ij}$ is the attention coefficient, $e_{ij}$, the attention score, $\mathbf{a}$ is a learnable weight vector, $\|$ denotes concatenation, and $\mathbf{W}$ transforms features into the attention space. Now, to handle the temporal sequence of $\mathbf{h}_i$, we used an LSTM layer which can handle the sequential traffic data, preserving important temporal relationships through specialized gating mechanisms. These gates control the flow of information within the LSTM cell as described below:
\begin{align}
    \mathbf{f}_t &= \sigma\left(\mathbf{W}_f \mathbf{x}_t + \mathbf{U}_f \mathbf{h}_{t-1} + \mathbf{b}_f\right) \\
    \mathbf{i}_t &= \sigma\left(\mathbf{W}_i \mathbf{x}_t + \mathbf{U}_i \mathbf{h}_{t-1} + \mathbf{b}_i\right) \\
    \mathbf{o}_t &= \sigma\left(\mathbf{W}_o \mathbf{x}_t + \mathbf{U}_o \mathbf{h}_{t-1} + \mathbf{b}_o\right) \\
    \tilde{\mathbf{c}}_t &= \tanh\left(\mathbf{W}_c \mathbf{x}_t + \mathbf{U}_c \mathbf{h}_{t-1} + \mathbf{b}_c\right) \\
    \mathbf{c}_t &= \mathbf{f}_t \odot \mathbf{c}_{t-1} + \mathbf{i}_t \odot \tilde{\mathbf{c}}_t \\
    \mathbf{h}_t &= \mathbf{o}_t \odot \tanh\left(\mathbf{c}_t\right)
\end{align}
where $\sigma$ is the sigmoid activation, $\mathbf{f}_t$, $\mathbf{i}_t$, and $\mathbf{o}_t$ are forget, input, and output gates, respectively, $\odot$ denotes element-wise multiplication, $\mathbf{x}_{t}$ is the concatenated node embedding at time \(t\), $\mathbf{c}_t$ is the cell state. Finally, the LSTM output $\mathbf{h}_t$ is then passed through a dense layer for prediction:
\begin{equation}
    \hat{\mathbf{F}}_{t+1:t+T} = \text{Dense}(\mathbf{h}_t)
\end{equation}
where $\hat{\mathbf{F}}_{t+1:t+T} \in \mathbb{R}^{|\mathcal{V}| \times T}$ represents the predicted flows for all nodes over the forecast horizon.

To measure the uncertainty in our forecasts and provide prediction intervals with coverage guarantees, we employ ACP. At each time step, the residuals between the predicted and actual traffic flows are calculated as:

\begin{equation}
r_t = \bigl|\,\hat{f}_t - f_t\,\bigr|
\end{equation}
where $f_t$ is the observed traffic flow, and $\hat{f}_t$ is its point prediction. The residuals on the validation set form the calibration set \(\mathcal{R}_{\text{cal}}\). For a desired coverage level \(1-\alpha\) (e.g.\ \(\alpha=0.10\) for 90\% confidence) we take the empirical \((1-\alpha)\)-quantile of this set:

\begin{equation}
q_{1-\alpha} \;=\;
\operatorname{Quantile}_{1-\alpha}\bigl(\mathcal{R}_{\text{cal}}\bigr)
\end{equation}%
which is the smallest value such that \(\Pr\{\,r_t \le q_{1-\alpha}\,\}\ge 1-\alpha\). Prediction intervals are then obtained by adding and subtracting \(q_{1-\alpha}\) from the point forecast:

\begin{equation}
\hat{F}_t - q_{1-\alpha} \;\le\; F_t \;\le\;
\hat{F}_t + q_{1-\alpha},
\end{equation}%
This method guarantees that the true flow \(F_t\) falls inside the interval with probability \(1-\alpha\). After each epoch, we fix the updated model, recompute \(q_{1-\alpha}\) on the unchanged validation set, and thus keep coverage stable as traffic patterns shift. Because the validation data are never used for weight updates, calibration remains strictly post-hoc, consistent with conformal-prediction theory. The overall workflow of the model is depicted in Algorithm 2.

\begin{algorithm}[H]
\caption{Traffic Flow Prediction with GAT-LSTM-ACP}\label{alg:alg2}
\begin{algorithmic}
\State \textbf{Input:} $\mathbf{F}, \mathbf{T}_{\text{mean}}, \mathbf{A}_{\text{avail}}, \text{CV} \in \{0.1, 0.3, 0.5, 0.7, 1.0\})$
\State \textbf{Output:} $\hat{\mathbf{F}}, \text{PI}$
\For{each $\text{CV}$}
    \For{$m = 1$ to $50$}
        \State \textbf{Step 1: Adaptive Adjacency Matrix}
        \State $\sigma_{\text{ln}} = \sqrt{\ln(\text{CV}^2 + 1)}$, $\mu_{\text{ln}} = \ln(\mathbf{T}_{\text{mean}}) - \frac{\sigma_{\text{ln}}^2}{2}$
        \State $\mathbf{T}_{\text{dynamic}}[i,j] \sim \text{LogNormal}(\mu_{\text{ln}}, \sigma_{\text{ln}})$
        \State $T_{ij}^\text{adjusted} = T_{ij}^\text{dynamic} \cdot \big(1 + \sum_{k} \alpha_k \cdot \rho_{ij}^k\big)$
        \State $\mathbf{A}_{\text{dynamic}}[i,j] = \exp\left(-\frac{\left(T_{ij}^\text{adjusted} / T_{\max}\right)^2}{2\sigma^2}\right)$
        \State $\mathbf{A}_{\text{adaptive}} = \mathbf{A}_{\text{dynamic}} \cdot \mathbf{A}_{\text{avail}}$

        \State \textbf{Step 2: Graph Attention (GAT)}
        \State $\mathbf{z}_{ij} = \mathbf{a}^\top \left[ \mathbf{W}\mathbf{h}_i, \mathbf{W}\mathbf{h}_j \right]$, $e_{ij} = \text{LReLU}(\mathbf{z}_{ij})$
        \State $\alpha_{ij} = \text{softmax}_j(e_{ij}) \cdot \mathbf{A}_{\text{adaptive}}[i,j]$
        \State $\mathbf{h}_i^{(l+1)} = \sigma\Big(\sum_{j \in \mathcal{N}(i)} \alpha_{ij}\mathbf{W}\mathbf{h}_j\Big)$

        \State \textbf{Step 3: LSTM with Attention}
        \State $\alpha_t = \text{softmax}(\tanh(\mathbf{H}\mathbf{W}_{\text{att}} + \mathbf{b}_{\text{att}}))$
        \State $\mathbf{c} = \sum_t \alpha_t \mathbf{h}_t$, $\hat{\mathbf{F}} = \text{Dense}(\mathbf{c})$
    
        \State \textbf{Step 4: ACP}
        \State $q_{1-\alpha} = \text{Quantile}_{1-\alpha}(|\hat{f}_t - f_t|)$
        \State $\text{PI} = [\hat{\mathbf{F}} - q_{1-\alpha}, \hat{\mathbf{F}} + q_{1-\alpha}]$
    \EndFor
\EndFor\\
\Return $\hat{\mathbf{F}}, \text{PI}$
\end{algorithmic}
\end{algorithm}

\begin{figure}[h!] \centering \includegraphics[width=\linewidth]{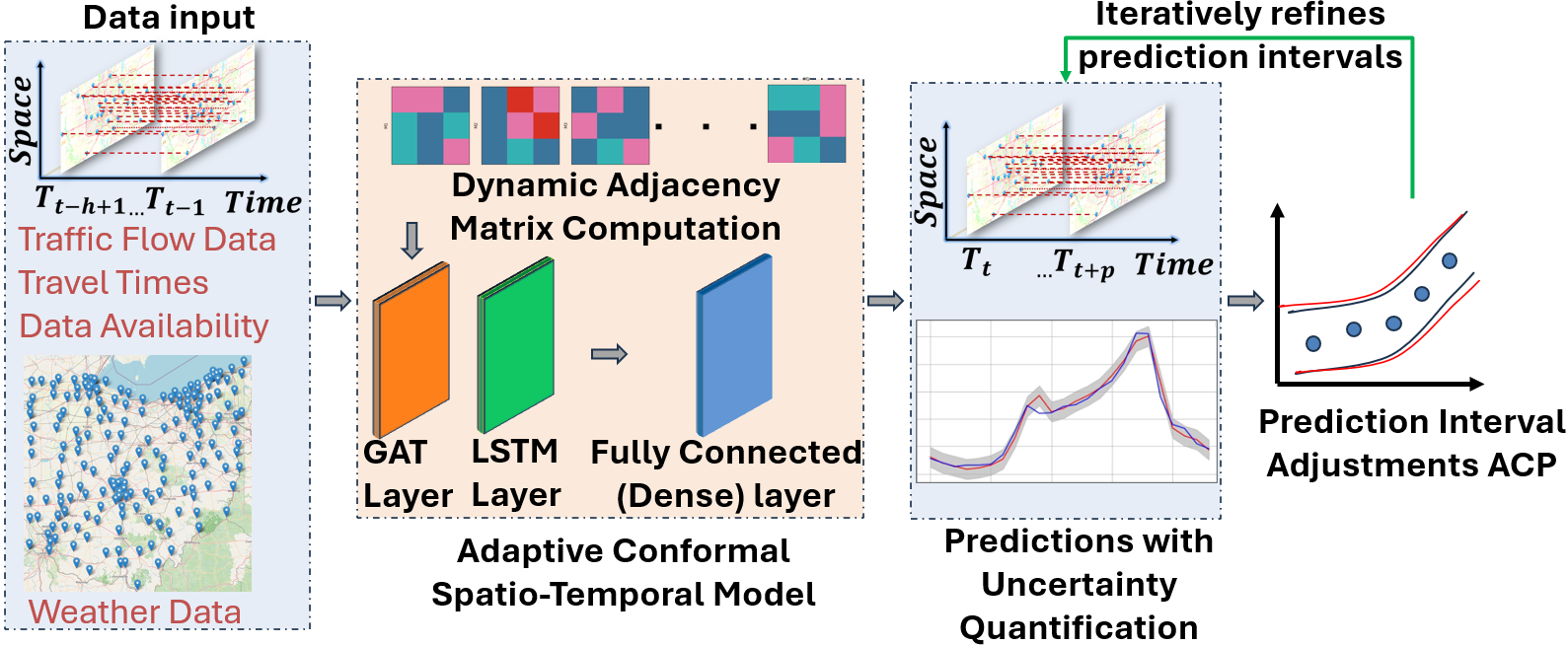} \caption{Model Architecture: Integration of GAT, LSTM, and ACP} \label{fig:model_architecture} \end{figure}

\noindent
Figure~\ref{fig:model_architecture} illustrates the model architecture, and interaction between the GAT layer, LSTM layer, and the ACP module.

\section{Results and Discussion}
\begin{figure}[h]
    \centering
    \includegraphics[width=\linewidth]{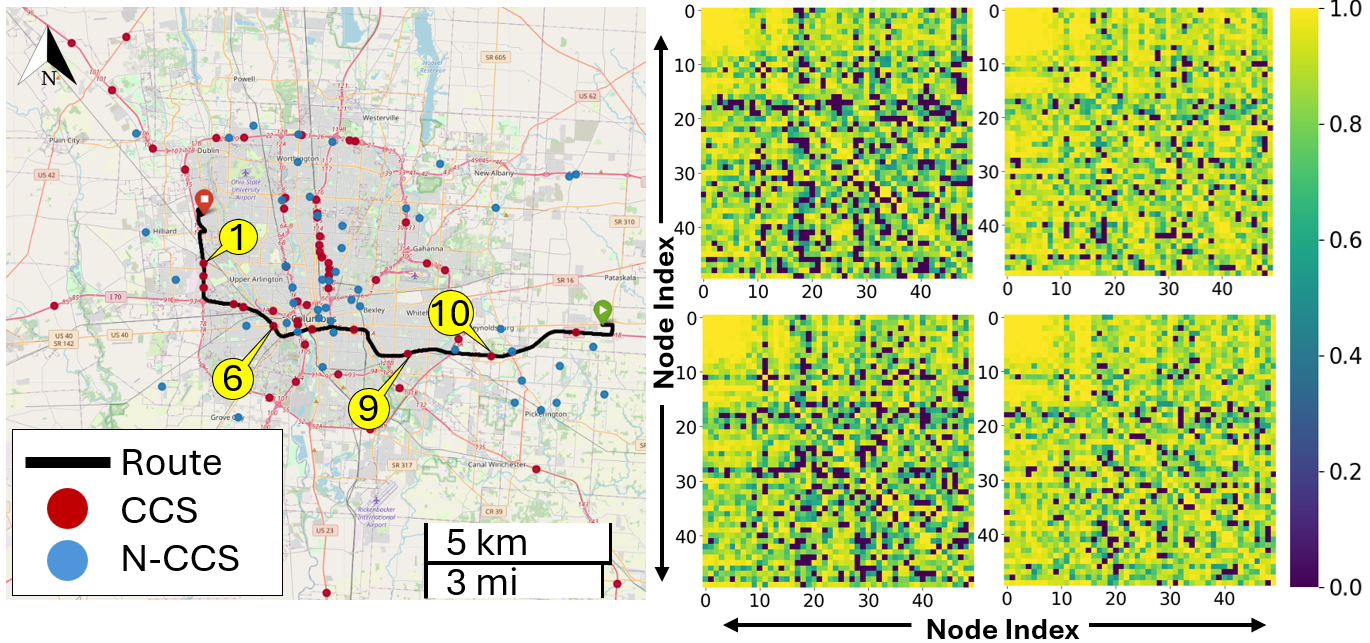} 
    \caption{Sampled Route (left); Example Adaptive Adjacency Matrices (right)}
    \label{fig:adj_matrices}
\end{figure}

We conducted experiments on an urban traffic route using 2019 ODOT-TCDS flow data and RWIS weather data for the nodes along the route in Columbus, OH. We ran five traffic scenarios with CV values of 0.1, 0.3, 0.5, 0.7, and 1.0. Figure \ref{fig:adj_matrices} shows the sample route and slices of adaptive adjacency matrices generated for these scenarios. The complete travel-time-based adjacency matrix consists of 273×273 node pairs.

\noindent
Table \ref{tab:weather_data} presents a sample weather dataset containing temperature, wind speed, and precipitation type collected at hourly intervals.
\vspace{-5pt}
\begin{table}[h]
\centering
\setlength{\tabcolsep}{2pt}
\renewcommand{\arraystretch}{1.2}
\caption{Sample Weather Dataset}
\label{tab:weather_data}
\begin{tabular}{|c|p{2cm}|p{1cm}|p{1.5cm}|p{2cm}|}
\hline
\textbf{ID} & \textbf{Date/Time (hourly)} & \textbf{Temp. (F)} & \textbf{Wind Speed (mph)} & \textbf{Precipitation Type} \\ \hline
200002 & 1/28/2019 00:00 & 44 & 9 & No Precip \\ \hline
200003 & 1/19/2019 18:00 & 33 & 12 & Heavy Rain \\ \hline
\multicolumn{5}{|c|}{\textbf{...}} \\ \hline
200185 & 1/20/2019 08:00 & 28 & 15 & Heavy Snow \\ \hline
\end{tabular}
\end{table}
\vspace{-10pt}

\subsection{Baseline Methods and Evaluation Metric}

For comparative analysis, we selected the following baseline algorithms:

\begin{itemize}
    \item Historical Average (HA) \cite{HA}: Uses average historical data for predictions.
    \item Autoregressive Integrated Moving Average (ARIMA) \cite{ARIMA}: A traditional statistical time-series forecasting method.
    \item Feedforward Neural Network (FNN): A simple neural network capturing nonlinear relationships.
    \item Long Short-Term Memory (LSTM) \cite{LSTM}: Models temporal dependencies in sequential data.
    \item Graph Convolutional Network (GCN) \cite{GCN}: Considers spatial dependencies using a static adjacency matrix.
    \item GCN-GRU \cite{Fu2016}: Combines spatial features via GCN and temporal modeling via GRU; uses the same inputs and adaptive adjacency as the proposed model.
    \item Store-and-Forward (SAF, link-queue) \cite{Aboudolas2009}: Physics-based discrete conservation with capacity.
    \item Link Transmission Model (LTM) \cite{Yperman07}: Cumulative kinematic-wave realization (LWR/Newell).

\end{itemize}
Since detector-level turning movements are unobserved, both SAF and LTM are constructed with proxy turning ratios from the travel-time matrix via an exponential impedance kernel and normalized over the top-k fastest neighbors \cite{Train2009, Fosgerau2013, Dial1971}. Also, delays are extracted from the time matrix, and per-station capacities are set from the 99th percentile of training counts. This approach is consistent with standard impedance-based route choice theory where turning rates are estimable \cite{Aboudolas2009}.

We evaluate prediction accuracy using Mean Absolute Errors (MAE) and Root Mean Squared Errors (RMSE), and uncertainty measure with mean prediction interval width (MPIW) and prediction interval coverage probability (PICP):

\begingroup                      
\setlength{\abovedisplayskip}{4pt}      
\setlength{\belowdisplayskip}{4pt}      
\setlength{\abovedisplayshortskip}{0pt} 
\setlength{\belowdisplayshortskip}{2pt}

\begin{equation}
\text{MAE} = \frac{1}{N} \sum_{i=1}^{N} \lvert F_i - \hat{F}_i \rvert
\end{equation}

\begin{equation}
\text{RMSE} = \sqrt{\frac{1}{N} \sum_{i=1}^{N} (F_i - \hat{F}_i)^2}
\end{equation}

\begin{equation}
\text{PICP} = \frac{1}{N} \sum_{i=1}^{N} \mathbf{1}\!\bigl(\hat{F}_i^L \le F_i \le \hat{F}_i^U\bigr)
\end{equation}

\begin{equation}
\text{MPIW} = \frac{1}{N} \sum_{i=1}^{N} \bigl(\hat{F}_i^U - \hat{F}_i^L\bigr)
\end{equation}
\endgroup                  

\noindent
where \(\hat{\mathbf{F}}_i^L\) and \(\hat{\mathbf{F}}_i^U\) are the lower and upper bounds of the prediction interval for the \(i\)-th observation, respectively. We omit MAPE due to its near-zero flow values, where division by small or zero values leads to undefined errors; MAE and RMSE remain more robust in such cases.

\subsection{Performance Comparison}

Table~\ref{tab:mae_rmse} presents the MAE and RMSE for each method across five CV scenarios. The prediction horizon was 15 minutes with a 24-hour historical look-back period. All models including the ablation variants were trained with early stopping (patience = 10) and a maximum of 20 epochs.

\begin{table}[!h]
\caption{Prediction Performance Comparison under Different Traffic Scenarios\label{tab:mae_rmse}}
\centering
\setlength{\tabcolsep}{4pt}
\renewcommand{\arraystretch}{1.1}
\begin{tabular}{|l|c|c|}
\hline
\multirow{2}{*}{\textbf{Method}} & \multicolumn{1}{c|}{\textbf{MAE}} & \multicolumn{1}{c|}{\textbf{RMSE}} \\
\cline{2-3}
& \textbf{0.1 / 0.3 / 0.5 / 0.7 / 1.0} & \textbf{0.1 / 0.3 / 0.5 / 0.7 / 1.0} \\
\hline
HA & 0.91/0.89/0.89/0.90/0.89 & 1.04/1.01/1.01/1.02/1.01 \\ \hline
FNN & 0.18/0.17/0.17/0.18/0.17 & 0.25/0.24/0.23/0.25/0.24 \\ \hline
LSTM & 0.20/0.18/0.17/0.20/0.19 & 0.27/0.26/0.24/0.27/0.28 \\ \hline
ARIMA & 0.14/0.14/0.15/0.15/0.15 & 1.45/1.47/1.49/1.48/1.47 \\ \hline
GCN & 0.17/0.16/0.17/0.18/0.16 & 0.25/0.26/0.31/0.29/0.30 \\ \hline
GCN-GRU & 0.15/0.14/\textbf{0.13}/\textbf{0.13}/0.16 & 0.24/0.22/\textbf{0.22}/\textbf{0.22}/0.25 \\ \hline
SAF & 0.59/0.63/0.62/0.61/0.58 & 0.87/0.94/0.92/0.91/0.86 \\ \hline
LTM & 0.81/0.80/0.82/0.81/0.82 & 1.17/1.17/1.19/1.16/1.17 \\ \hline
\textbf{Proposed Model} & \textbf{0.13}/\textbf{0.12}/0.14/0.14/\textbf{0.13} & \textbf{0.22}/\textbf{0.21}/0.23/0.23/\textbf{0.22} \\ \hline
\end{tabular}
\end{table}

\noindent
The proposed model consistently outperformed all baseline methods across most traffic scenarios in both MAE and RMSE, demonstrating strong robustness to traffic-flow variability. The only notable exception is the GCN-GRU baseline, which recorded the lowest error under the mid-range conditions of CV = 0.5 and 0.7. This result suggests that a lightweight graph–recurrent architecture can be effective when variability is moderate and the spatial structure remains relatively stable. The physics-based models (SAF and LTM) observed higher errors across all CVs due to their conservative capacity-delay dynamics and absence of feature learning. Outside the mid-range, the proposed model maintained better performance in the extreme regimes (CV = 0.1, 0.3, and 1.0) where either very low or very high fluctuations challenge static-graph models. The end-to-end training times for CV values of 0.1, 0.3, 0.5, 0.7, and 1.0 were 182, 230, 265, 212, and 199 minutes, respectively, on a system with an Intel i7-8750H CPU, 16 GB RAM, and an NVIDIA GTX 1050 GPU.

\subsection{Prediction Results}

\begin{figure}[h] \centering \includegraphics[width=\linewidth]{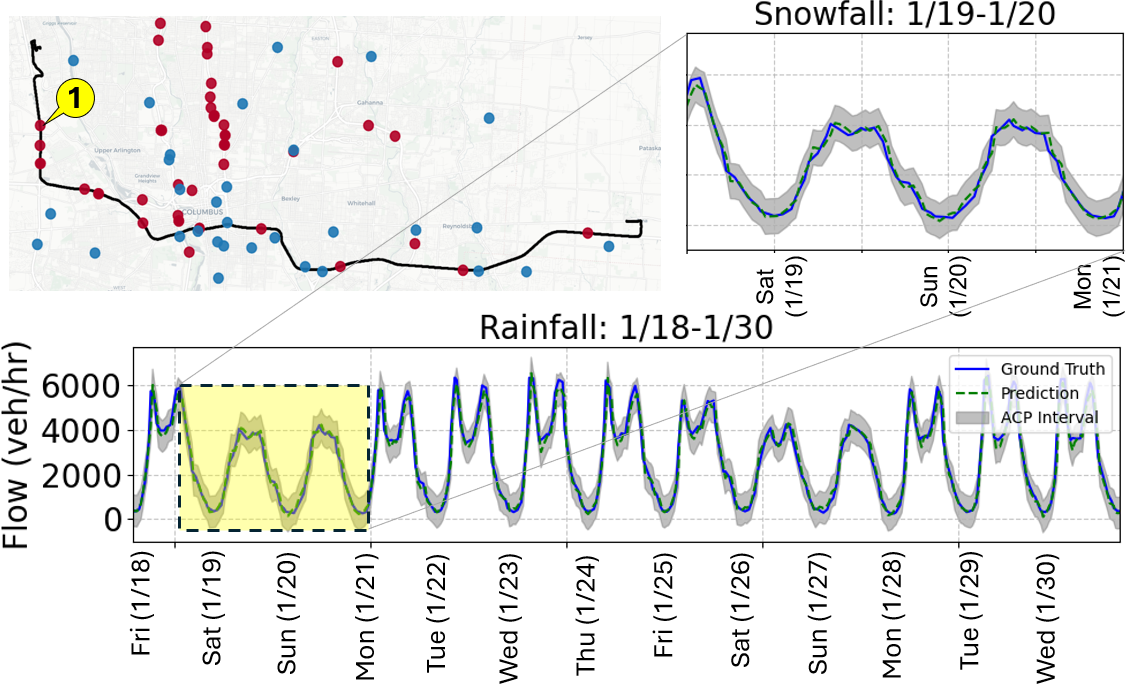} \caption{Traffic Prediction with Uncertainty Bounds for 2019 Rain and Snow} \label{fig:prediction_bounds} \end{figure}

Figure~\ref{fig:prediction_bounds} shows traffic flow prediction results with uncertainty bounds for node 1 for a sample CV = 0.5 scenario. The top-right plot highlights snowfall (1/19-1/20), and the bottom plot focuses on rainfall (1/18-1/30), demonstrating the model's adaptability to varying weather conditions while maintaining reliable predictions.

\subsection{Uncertainty Quantification}
For uncertainty quantification, we compared the proposed ACP method against the following baseline techniques:

\begin{itemize}
    \item Gaussian Process Regression (GPR) \cite{GPR}: Provides uncertainty through a probabilistic model.
    \item Mean-Variance Estimation (MVE) \cite{MVE}: Predicts both mean and variance for uncertainty estimation.
    \item Deep Ensembles (DE) \cite{DE}: Estimates uncertainty by variance calculation across multiple independently trained models.
    \item Quantile Regression (QR) \cite{QR}: Directly predicts conditional quantiles for confidence intervals.
    \item Bootstrap Aggregation (BA) \cite{BA}: Combines predictions from resampled datasets to estimate variability.
    \item Conformal Prediction (CP) \cite{CP}: Constructs intervals using residuals from a base model and a calibration set.
    \item Conformalized Quantile Regression (CQR) \cite{CQR}: Enhances quantile regression with conformal calibration for adaptive intervals.
\end{itemize}

\noindent
Table~\ref{tab:picp_mpiw} compares the PICP and MPIW of the ACP method against baseline uncertainty quantification techniques.

\begin{table}[h]
\caption{PICP and MPIW Comparison under Different Traffic Scenarios\label{tab:picp_mpiw}}
\centering
\setlength{\tabcolsep}{3pt} 
\renewcommand{\arraystretch}{1.1} 
\begin{tabular}{|l|c|c|}
\hline
\multirow{2}{*}{\textbf{Method}} & \multicolumn{1}{c|}{\textbf{PICP \%}} & \multicolumn{1}{c|}{\textbf{MPIW}} \\
\cline{2-3}
& \textbf{0.1 / 0.3 / 0.5 / 0.7 / 1.0} & \textbf{0.1 / 0.3 / 0.5 / 0.7 / 1.0} \\
\hline
GPR                   & 91.02/90.94/91.08/91.33/\textbf{92.32} & 0.86/0.88/0.92/0.98/1.03 \\ \hline
MVE                   & 90.01/89.57/89.85/89.24/88.07          & 0.44/0.45/0.47/0.46/0.42 \\ \hline
DE                    & 63.18/62.09/57.15/59.44/60.36          & 0.16/0.15/0.14/0.15/0.15 \\ \hline
QR                    & 90.45/89.93/88.22/90.08/91.16          & 0.43/0.42/0.40/0.41/0.39 \\ \hline
BA                    & 65.93/66.41/61.02/60.35/61.63          & 0.17/0.16/0.17/0.17/0.15 \\ \hline
CP                    & 89.72/90.11/90.46/90.35/90.87          & 0.70/0.73/0.84/0.81/0.77 \\ \hline
CQR                   & 91.13/90.56/91.01/91.47/91.25          & 0.66/0.76/0.82/0.79/0.80 \\ \hline
\textbf{Proposed ACP} & \textbf{92.21}/\textbf{91.46}/\textbf{92.76}/\textbf{91.62}/91.97 & 0.68/0.71/0.78/0.78/0.77 \\
\hline
\end{tabular}
\end{table}

\noindent
The proposed ACP method maintained competitive PICP across all CV scenarios, consistently delivering reliable and adaptive uncertainty intervals. While methods like QR achieved narrower MPIW, it achieves it at the expense of lower coverage and they do not adjust as conditions change (PICP = 88.22\%), potentially underestimating uncertainty in dynamic traffic conditions. Whereas, ACP maintained high coverage (PICP = 92.76\%) with moderately wider intervals (MPIW = 0.78), achieving a trade-off between interval sharpness and coverage reliability. Since uncertainty quantification aims to ensure that the true values fall within the predicted intervals with high probability, slightly wider intervals are acceptable if they lead to more reliable coverage. In traffic forecasting scenarios, under-coverage poses a greater risk than marginally wider intervals, as missing true observations can lead to unsafe decisions. Therefore, despite higher MPIW, ACP offers a practical advantage by integrating calibration into the training loop rather than assuming a fixed uncertainty profile.

\subsection{Ablation Study}
To assess the impact of individual components, we conducted an ablation study to see how adaptive adjacency with weather integration affects the model performance. We decided to test the following configurations: 
(1) Adaptive Adjacency matrix with weather (AAW-Base), (2) decreased LSTM width from H=64 to H=32 (AAW‑rLSTM), (3) increased attention heads from h=4 to h=8 (AAW‑iGAT), (4) Replaced dynamic adjacency with a static adjacency matrix (StatAdj-W), and (5) Removed both dynamic adjacency and weather (StatAdj). The results are shown in Table~\ref{tab:ablation_study}.

\begin{table}[h]
\caption{Ablation Study under Different Traffic Scenarios\label{tab:ablation_study}}
\centering
\setlength{\tabcolsep}{6pt}
\renewcommand{\arraystretch}{1.1}
\begin{tabular}{|l|c|c|}
\hline
\multirow{2}{*}{\textbf{Configuration}} &
\multicolumn{1}{c|}{\textbf{MAE}} &
\multicolumn{1}{c|}{\textbf{RMSE}} \\
\cline{2-3}
& \textbf{0.1 / 0.3 / 0.5 / 0.7 / 1.0}
& \textbf{0.1 / 0.3 / 0.5 / 0.7 / 1.0} \\
\hline
\textbf{AAW‑Base} & \textbf{0.13}/\textbf{0.12}/\textbf{0.14}/\textbf{0.14}/\textbf{0.13} &
\textbf{0.22}/\textbf{0.21}/\textbf{0.23}/\textbf{0.23}/\textbf{0.22} \\ \hline
AAW‑rLSTM           & 0.14/0.14/0.15/0.15/0.16 & 0.23/0.23/0.24/0.24/0.26 \\ \hline
AAW‑iGAT            & 0.13/0.14/0.14/0.15/0.15 & 0.22/0.23/0.23/0.24/0.24 \\ \hline
StatAdj-W               & 0.14/0.15/0.16/0.17/0.16 & 0.23/0.24/0.26/0.28/0.26 \\ \hline
StatAdj               & 0.23/0.23/0.25/0.24/0.25 & 0.38/0.39/0.39/0.40/0.41 \\
\hline
\end{tabular}
\end{table}

\noindent
The findings show that AAW-Base achieved the lowest errors across all CVs (MAE: 0.12-0.14, RMSE: 0.21-0.23). AAW-rLSTM trained faster but increased errors by 0.01-0.03. AAW-iGAT matched AAW-Base for CV = 0.1 and 0.5, but showed no gains at higher CVs and increased training time by $\sim27\%$. StatAdj-W, without dynamic adjacency, increased errors further, while StatAdj, removing both dynamic adjacency and weather, performed worst (MAE up to 0.25, RMSE up to 0.41). Overall, combining adaptive adjacency and weather gave the best results.

\subsection{Model Validation with Simulation}

Going one step further, we validate the model's prediction by conducting SUMO traffic simulations. We extracted the predicted values for the nodes along the same route chosen in our experiment and then, created traffic scenarios following the methodology in \cite{PatilSUMO}. The route characteristics are outlined in Table~\ref{tab:route_characteristics}.

\begin{table}[h]
\caption{Route Characteristics\label{tab:route_characteristics}}
\centering
\setlength{\tabcolsep}{10pt}
\begin{tabular}{|l|c|}
\hline
\textbf{Parameter} & \textbf{Value} \\
\hline
Route Length (mi) & 32.43 \\
\hline
Number of Intersections & 19 \\
\hline
Number of Traffic Lights & 6 \\
\hline
Peak Volume (vehicles/hour) & 19,860 \\
\hline
Mean Speed (mph) & 56.16 \\
\hline
Number of Lanes & 2--4 \\
\hline
Functional Class & Local, Minor/Principal Arterials \\
\hline
\end{tabular}
\vspace{-10pt}
\end{table}

\begin{figure}[!h]
    \centering
    \includegraphics[width=\linewidth]{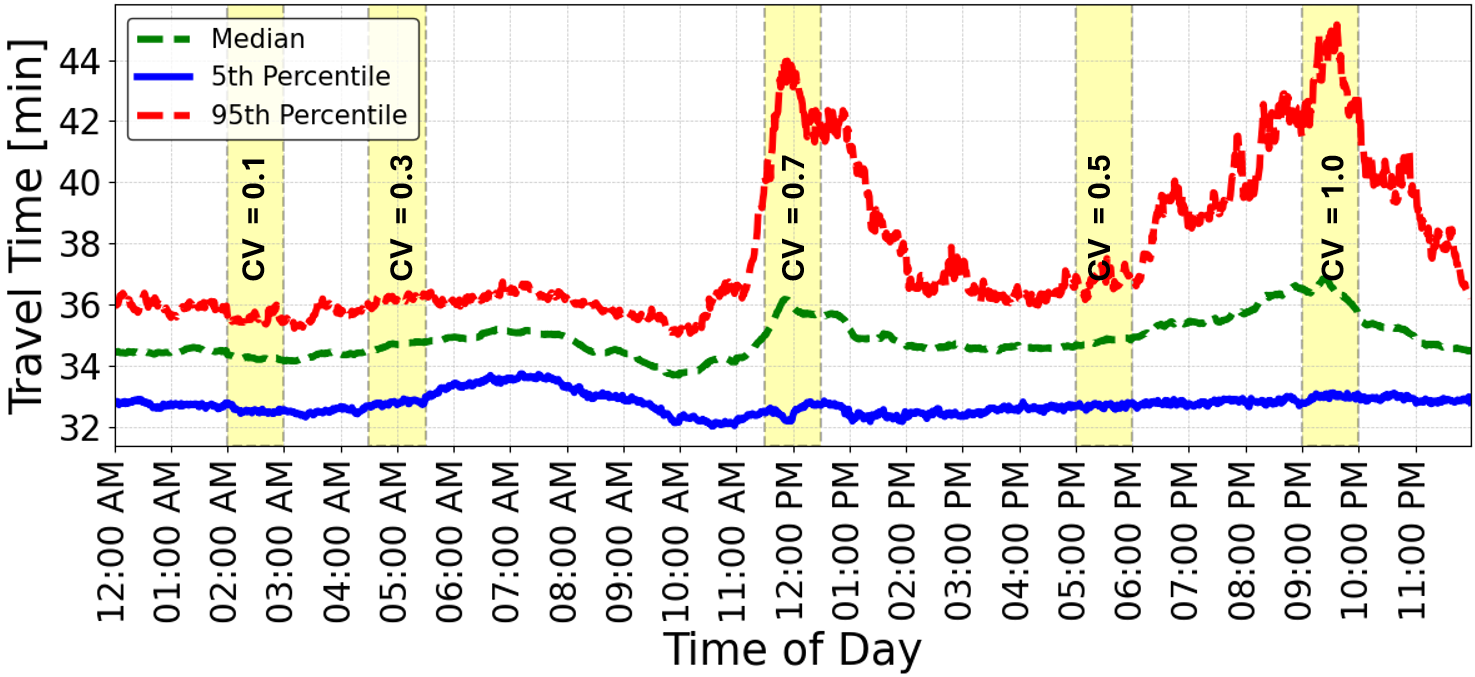}
    \caption{INRIX historical travel time data; shaded windows mark representative CV regimes}
    \label{fig:inrix_cv_windows}
\end{figure}

\noindent
We use the INRIX travel-time data only to select time windows for each CV level; for those hours, we take our model’s predicted node flows and feed them into SUMO. Figure~\ref{fig:inrix_cv_windows} marks the chosen windows. We run a 200-trial Monte Carlo study in two regimes to show how variability scales with CV\,=\,0.5 (medium, 3{:}30-4{:}30\,PM) and CV\,=\,1.0 (high variability, evening peak, 9{:}00-10{:}00\,PM). Please note that the shaded windows are illustrative.

\begin{figure}[!h] \centering \includegraphics[width=\linewidth]{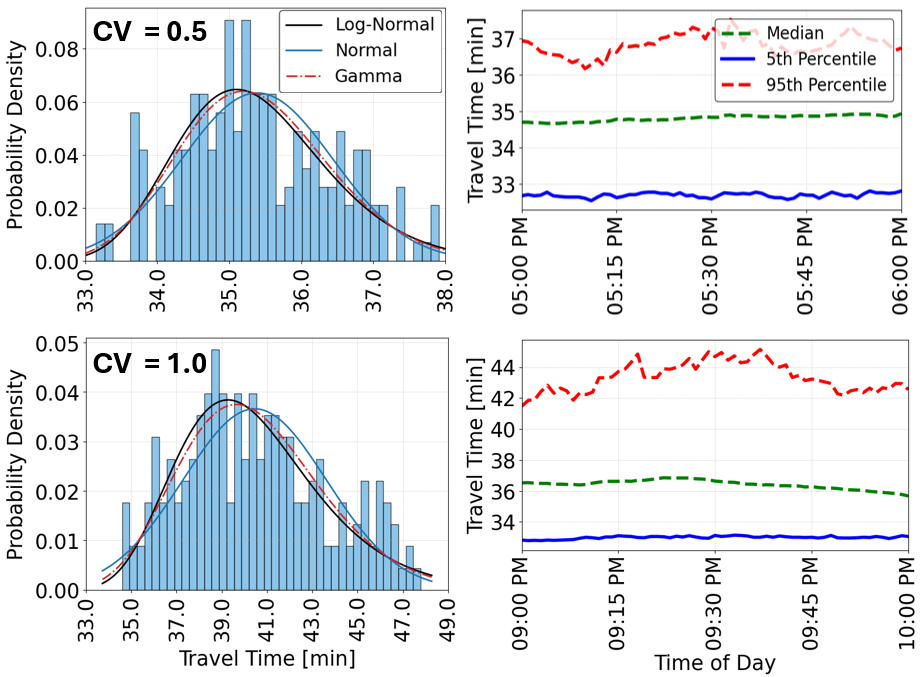} \caption{Comparison of simulated and INRIX travel time distributions} \label{fig:validation} \end{figure}

The resulting SUMO VUT travel-time samples in Figure~\ref{fig:validation} follow the log-normal form assumed in Equation 6, with mean times of 35.39 and 40.46 minutes, respectively.

Kolmogorov–Smirnov (KS) tests confirmed log-normal distribution: 
\begin{itemize}
  \item CV = 0.5: log-normal (KS = 0.042/p = 0.91), normal (KS = 0.084/p = 0.11), gamma (KS = 0.063/p = 0.61)
  \item CV = 1.0: log-normal (KS = 0.036/p = 0.94), normal (KS = 0.068/p = 0.29), gamma (KS = 0.052/p = 0.79)
\end{itemize}
Lower KS and \(p{>}0.05\) for log-normal validate both the assumed distribution and simulation fidelity.

As expected, the higher CV case produces a wider travel-time distribution, while the medium CV case is tighter because congestion and demand fluctuations increases small disturbances during peaks, creating stop-and-go waves and more variable speeds; by contrast, lighter conditions stay near free-flow, so times cluster closely around the mean. This study supports our CV-driven, log-normal sampling for building the travel-time-based adaptive adjacency matrix.

\section{Conclusion} In this paper, we presented an urban traffic flow prediction framework that integrates adaptive adjacency matrices based on real-world travel time variability and weather conditions into a GNN-based model. We demonstrated how to dynamically account for traffic variability and environmental factors, achieving significant accuracy improvements compared to baseline methods, along with robust uncertainty quantification using ACP. Our model was validated through Monte-Carlo SUMO simulations, showing strong agreement with INRIX data. However, computational complexity leads to longer training times, which can be alleviated by using high-performance computing resources. Additionally, reliance on travel time and weather data limits applicability in areas with sparse sensor coverage. Another practical challenge is the dependence on low-latency sensor feeds, as ODOT stations report at varying intervals and may experience delays, potentially causing data gaps. Refreshing the graph and recalibrating ACP each minute can also take several seconds on standard servers, delaying real-time forecasts. Lastly, if this framework has to integrate into ODOT’s existing systems, it will require addressing differences in data formats to align outputs with existing dashboards and message signs. On that account, future efforts will focus on improving scalability and generalizability by enhancing computational efficiency, expanding datasets, and incorporating additional factors like road incidents and construction events. Overall, the proposed model demonstrates strong potential to enhance intelligent transportation systems and real-time traffic management.

 
\vspace{-30pt}

\end{document}